# Integrating automatic speech recognition into remote healthcare interpreting: A pilot study of its impact on interpreting quality

**Shiyi Tan**
University of Surrey, UK
s.tan@surrey.ac.uk

**Constantin Orăsan**
University of Surrey, UK
c.orasan@surrey.ac.uk

**Sabine Braun**
University of Surrey, UK
s.braun@surrey.ac.uk

## Abstract

This paper reports on the results from a pilot study investigating the impact of automatic speech recognition (ASR) technology on interpreting quality in remote healthcare interpreting settings. Employing a within-subjects experiment design with four randomised conditions, this study utilises scripted medical consultations to simulate dialogue interpreting tasks. It involves four trainee interpreters with a language combination of Chinese and English. It also gathers participants' experience and perceptions of ASR support through cued retrospective reports and semi-structured interviews. Preliminary data suggest that the availability of ASR, specifically the access to full ASR transcripts and to ChatGPT-generated summaries based on ASR, effectively improved interpreting quality. Varying types of ASR output had different impacts on the distribution of interpreting error types. Participants reported similar interactive experiences with the technology, expressing their preference for full ASR transcripts. This pilot study shows encouraging results of applying ASR to dialogue-based healthcare interpreting and offers insights into the optimal ways to present ASR output to enhance interpreter experience and performance. However, it should be emphasised that the main purpose of this study was to validate the methodology and that further research with a larger sample size is necessary to confirm these findings.

## 1 Introduction

Since the introduction of simultaneous interpreting (SI) through electro-acoustic sound transmission systems, technological advances have continuously shaped the world of interpreting. They have given rise to new forms of interpreting, including technology-mediated interpreting or distance interpreting, technology-supported interpreting or computer-assisted interpreting (CAI), and even technology-generated interpreting or machine interpreting (Braun, 2019).

Currently, one of the most promising technologies used to support interpreting workflows is automatic speech recognition (ASR) which converts human speech signals into a sequence of words using computer programmes (Jurafsky and Martin, 2008). In the context of interpreting, ASR was initially utilised to automate the process of querying glossaries and retrieving information in CAI tools (Fantinuoli, 2017). Driven by classic machine learning technologies including Hidden Markov Models and more recently by deep learning technologies, ASR now shows increasingly robust performance and can directly support the interpreting process by providing real-time transcripts of source speeches. Researchers were thus motivated to explore the practical application of ASR in interpreting. Improved accuracy for the rendition of "problem triggers" such as numbers, specialised terms and proper names in SI was achieved with ASR output (e.g., Desmet *et al.*, 2018; Defrancq and Fantinuoli, 2021). This initial focus on ASR's role in addressing specific stumbling blocks resulted in little attention being paid to its impact on overall interpreting quality. Few studies put the spotlight on ASR in consecutive interpreting (CI) as interpreters often can rely on notetaking as a memory aid and face less time pressure in CI compared to SI.

ASR output generally is neither entirely error-free, nor fully synchronised with the acoustic signal, possibly causing inaccuracies, delays and distractions for interpreters. Consequently, the investigation of how ASR can impact interpreting quality as a whole, is equally important for its wider adoption and development. The integration of ASR into CI-based public service interpreting also deserves attention, particularly in the contexts of healthcare and legal interpreting where accuracy can be a matter of "life and death". These types of interpreting tasks often feature obscure terminology and frequent use of numbers, units and dates which require correct rendition.

As an extra source of information, ASR output also competes for interpreters' cognitive resources with other processing tasks during interpreting such as comprehension and production (Seeber, 2011). The number of studies that have explored this problem is limited (e.g., Yuan and Wang, 2024; Li and Chimel, 2024), largely due to the complexity and challenges involved in examining cognitive performance.

The current study, as a pilot study for a larger research project, explores the integration of ASR into dialogue-based healthcare interpreting in an attempt to understand whether ASR helps or hinders interpreters. The rest of the paper is structured as follows. We begin by revisiting previous studies on the intersection of ASR and interpreting, followed by a description of the methodology in Section 3. Section 4 presents the results of the pilot study and Section 5 is a discussion of preliminary findings. The paper concludes with a summary of insights and limitations as well as an outline for future work.

## 2    Literature review

This section reviews previous explorations into ASR in interpreting from two aspects: the assessment of ASR systems' performance in interpreting and its impact on interpreting quality. Following a summary of research gaps, three research questions are proposed.

**Assessment of ASR systems' performance in interpreting**

Certain criteria need to be met by ASR systems to be applied in interpreting. An ASR system should be speaker-independent, able to manage continuous speech, support large-vocabulary recognition and provide the option to add specialised terms for improved recognition (Fantinuoli, 2017). A low word error rate (WER), a metric measuring transcription errors, and a low real-time factor (RTF), a metric assessing transcription speed are also expected in ASR systems (Fantinuoli, 2017).

Using three English texts containing 119 terms and 11 numerals, Fantinuoli (2017) tested that Dragon Naturally Speaking, an ASR engine integrated into the CAI tool InterpretBank 4, reached an accuracy of approximately 95% for term transcription after importing a list of English specialised terms from a bilingual glossary and 100% for numeral transcription. Student interpreters can maintain accuracy and fluency in SI with a 3-second latency in an automatic suggestion system for numbers (Fantinuoli and Montecchio, 2022). Using the Google Cloud Speech-to-Text API for ASR, InterpretBank demonstrated low latency and high precision (96%) in number transcription (Defrancq and Fantinuoli, 2021). In Fritella's study (2022), the latency of SmarTerp, an ASR-integrated CAI tool, was 2 seconds in transcribing name entities, specialistic terms and numbers.

In relation to the format of the transcribed text, research by Defrancq and Fantinuoli (2021) noted that the running transcript was a distraction for some students and preferences were divided regarding what aspects of figures to be displayed, such as only numbers or both numbers and units, and how they should be displayed on the screen.

**Impact of ASR on interpreting quality**

Currently, most research has investigated the impact of ASR on the rendition of numbers and specialised terms. As a result of displaying the numbers on slides, the accuracy rate of number interpreting rose from 56.5% to 86.5% (Desmet *et al.*, 2018). Defrancq and Fantinuoli (2021) found that the interpreting accuracy rates of nearly all number types were enhanced when ASR was available, a finding echoed by Pisani and Fantinuoli (2021), who reported a significant decline in the error rate of number renditions. A difference between the two studies lies in the way the transcribed numbers were presented. Numbers were embedded and highlighted in the entire transcript in the former study, while in the latter, numbers were shown in isolation. With Zoom live captioning, the error rates in interpreting interest periods containing numbers and proper names saw a 30% reduction (Yuan and Wang, 2023).

To date, only a few studies have examined the effectiveness of ASR in relation to overall interpreting quality with various quality assessment frameworks being adopted. Cheung and Li (2022) found that the presence of captions in a video enhanced accuracy but reduced fluency among student interpreters, based on two scoring sheets for each measure. A significant improvement in overall interpreting performance with live captions was also observed among trainees, using quality assessment criteria from the researchers' institution (Yuan and Wang, 2024). In an experiment with professional interpreters, Rodríguez González *et al.* (2023) reported a notable decline in the total number of interpreting errors with ASR support, although style-related errors increased, as assessed through the NTR model (Romero-Fresco and Pöchhacker, 2017). However, all these studies pertained to simultaneous interpreting.

In relation to consecutive interpreting, Chen and Kruger (2022) introduced a computer-assisted consecutive interpreting (CACI) mode that integrates ASR technology with machine translation (MT). Different from studies that employed ASR as a supplementary tool during interpreting, this study required interpreters to listen to the source speech and respeak it into iFLYTEK, an ASR system generating textual output, which was subsequently translated by an MT system. The interpreters then produced a target speech by consulting both the ASR-generated text and the MT output. In CACI, overall Chinese-to-English interpreting quality was enhanced, and fluency was improved in both directions. With a similar research design, Wang and Wang (2019) found that the accuracy of CI was enhanced with ASR-supported MT reference being provided, but no clear conclusion was reached regarding fluency. However, it is important to be aware that in these studies, the differences observed resulted from the combined effects of ASR and MT, making it impossible to draw conclusions about ASR alone.

**Research gaps in ASR-supported interpreting**

Several research gaps have emerged from the reviewed studies. First, as most previous studies focused solely on using ASR to support the interpreting of "problem triggers", such as numbers and terms, the impact of ASR on overall interpreting quality remains underexplored.

Second, most ASR systems used are off-the-shelf software, leaving little leeway to adjust transcription accuracy or customise output format in experiments. Generally, these ASR systems can be classified into four types (Table 1).

Third, diversity was observed in the presentation of ASR-generated text, ranging from only numbers (e.g., Fantinuoli and Montecchio, 2022; Desmet *et al.*, 2018) to entire transcripts with or without numbers being highlighted (e.g., Defrancq and Fantinuoli, 2021; Rodríguez González *et al.*, 2023; Saeed *et al.*, 2023), from chunked segments (Cheung and Li, 2022) to scrolling captions (Yuan and Wang, 2023). Some studies divided the interface into distinct sections to display different types of transcribed text, such as numbers with units of measurement, proper names and specialised terms (Fantinuoli *et al.*, 2022), named entities, terms and numbers (Frittella, 2022), and terminology and numerals (Fantinuoli, 2017). With these variations, no agreement was reached on the optimal way of presenting ASR output, and no study tested varying types of ASR output.

Fourth, most of the reviewed research engaged student interpreters, with only a few studies involving professional interpreters (e.g., Frittella, 2022; Rodríguez González *et al.*, 2023; Li and Chmiel, 2024). Although trainee interpreters are more accessible than experienced interpreters for experimental and pedagogical purposes, the significance of involving professional interpreters is crucial, especially concerning the application of ASR in authentic tasks.

| Types of ASR systems | Specific tools and Key studies |
|---|---|
| Simulated ASR systems | Slides (Desmet *et al.*, 2018) |
|  | Video (Fantinuoli and Montecchio, 2022) |
| CAI tools with ASR features | InterpretBank (Fantinuoli, 2017; Defrancq and Fantinuoli, 2021; Pisani and Fantinuoli, 2021) |
|  | SmarTerp (Frittella, 2022) |
|  | KUDO Interpreter Assist (Fantinuoli *et al.,* 2022) |
| Stand-alone ASR engines | iFLYTEK (Chen and Kruger, 2022) |
|  | Dragon Anywhere (Wang and Wang, 2019) |
| Platforms with captioning features | Zoom captioning (Yuan and Wang, 2023) |
|  | YouTube subtitles (Li and Chmiel, 2024) |

Table 1: ASR systems used in previous studies

Last, current studies also vary in two research design-related factors that presumably impact results: whether participants received ASR training prior to the experiments and which quality assessment framework was implemented.

Given these gaps, we proposed three research questions for the full research project:

1) Is there a significant difference in overall interpreting quality between interpreting with varying types of ASR support and without ASR support?

2) Does the interpreting quality vary across different types of ASR output?

3) How do interpreters interact with different types of ASR output?

In a mixed-methods approach, we conducted experiments complemented by post-experiment retrospective reports and semi-structured interviews in a pilot study, to tentatively explore answers to these questions.

## 3 Methodology

This section describes the participant information, interpreting materials, experiment design and procedure as well as data analysis methods.

**Participants**

In the pilot study, four trainee interpreters (all females, mean age = 27.5, range = 24-31, $SD$ = 3.51) were recruited within the guidelines of the ethics committee. They were recent graduates from a one-year master's programme in interpreting at a university in England, where they all had completed four compulsory modules on CI and SI. Three of them held a bachelor's degree in English or Translation. All spoke Chinese as their mother tongue and English as their second language. Their average IELTS score was 7.0 (range = 6.5-7.5, $SD$ = 0.41). Their prior use of ASR software was limited to classroom demonstrations.

**Materials**

The interpreting materials used in this study were four scripts adapted from authentic medical consultations provided by a private hospital in London. The scripts are four consecutive consultations between a nephrologist and a patient with renal disease. The difficulty of the four scripts (Table 2) was controlled to ensure comparability based on word count, duration, speed and the Flesch reading ease index, which measures how difficult a text is to understand. A score between 60 and 70 indicates that the text is written at a standard level of readability and can be easily understood by individuals aged 13 to 15. The scripts were recorded into videos by an English native speaker portraying the doctor and a Chinese native speaker acting as the patient.

The Microsoft Azure Speech Service API (Microsoft) was called to generate bilingual transcripts of the consultations, chosen for its relatively high accuracy and low latency. No domain customisation was applied. The word error rates of the four scripts ranged from approximately 15% to 20%.

|  | Word count (words) | Duration (minutes) | Speed (wpm) | Flesch reading ease index | Word Error Rate (English utterances) |
|---|---|---|---|---|---|
| **Script 1** | 756 | 6'07" | 124 | 70.2 | 19.80% |
| **Script 2** | 772 | 5'28" | 141 | 66.1 | 14.62% |
| **Script 3** | 742 | 5'33" | 134 | 62.0 | 17.04% |
| **Script 4** | 779 | 5'55" | 132 | 61.3 | 19.68% |

Table 2: Difficulty control and word error rate of the scripts

**Apparatus**

To provide different types of ASR output, this study opted not to use CAI tools with ASR functions or platforms with captioning features. An interface (Figure 1) was designed by the research team for interpreters to carry out video remote interpreting tasks under various conditions. In the top left, a video player is displayed, while the right side features an ASR section with three text boxes. The current utterance appears in the bottom text box and gradually moves up as the next utterance is transcribed. The transcript related to an utterance was programmed to automatically appear immediately upon completion of the utterance. At that point the video was automatically paused to enable the interpreter to interpret. A blue "Next" button at the bottom allows the interpreter to listen to the next utterance either by pressing the spacebar on the keyboard or by clicking the mouse.

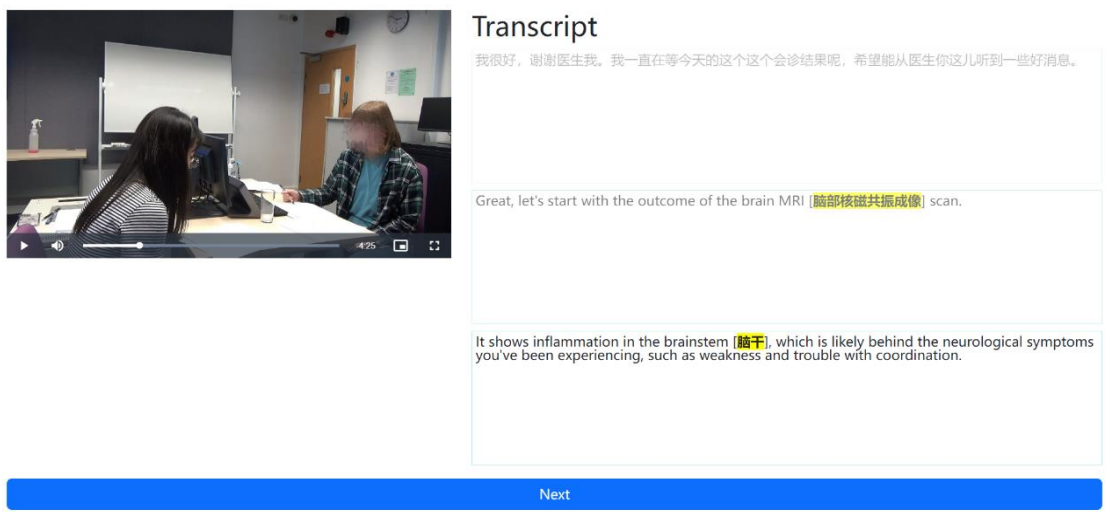

Figure 1: Interface for remote video interpreting with ASR

## Experiment design

To test how various types of ASR output affect interpreting quality, a baseline condition without ASR support was first devised, followed by three conditions with different types of ASR support. The conditions are as follows (Figure 2):

**Condition 1**: interpreting without ASR support

**Condition 2**: interpreting with partial ASR support (including the transcription of specialised terms and numbers and their translations)

**Condition 3**: interpreting with full ASR support (including the transcription of entire dialogue with the translations of numbers and specialised terms)

**Condition 4**: interpreting with ASR-fed ChatGPT summary (including a bullet-point summary with the translations of numbers and specialised terms). In this condition we provided ChatGPT with the following prompt to generate summaries:

*There is the output of an ASR system which transcribed a doctor-patient conversation.*
*The output is used by an interpreter to support their interpretation.*
*Shorten the output to about half length making sure that the important information is kept.*
*Make sure you keep the important information.*
*This short version will be shown to the interpreter to help them interpret the conversation.*
*Present the output using bullet points.*

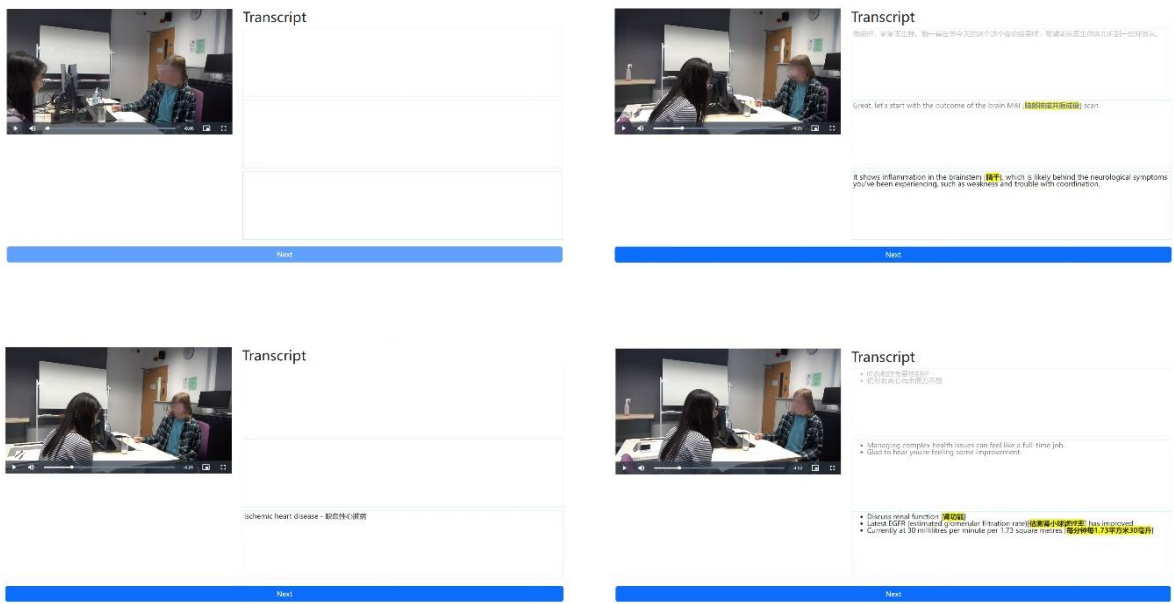

Figure 2: Interface without ASR support (top left), interface with partial ASR support (bottom left), interface with full ASR support (top right) and interface with ASR-fed ChatGPT summary (bottom right)

|  | Task 1 | Task 2 | Task 3 | Task 4 |
|---|---|---|---|---|
| **Participant 1** | Condition 1 | Condition 2 | Condition 3 | Condition 4 |
| **Participant 2** | Condition 2 | Condition 3 | Condition 4 | Condition 1 |
| **Participant 3** | Condition 3 | Condition 4 | Condition 1 | Condition 2 |
| **Participant 4** | Condition 4 | Condition 1 | Condition 2 | Condition 3 |

Table 3: Randomised experiment design

Table 3 shows a randomised Latin square experiment design to ensure that each interpreter experiences every condition equally.

**Procedure**

This pilot study was conducted from 5th to 9th August 2024. Before the experiment, all participants completed a pre-experiment questionnaire concerning demographic information, English proficiency, interpreting experience, prior experience with using ASR for interpreting tasks and vision conditions via Qualtrics.

On the experiment day, participants came to our interpreting lab. They were first briefed on the theme of the consultations and the names of the interlocutors. To closely simulate authentic medical interpreting, they were allotted only 15 minutes to prepare. Before each task, they received a 9-point eye calibration to ensure accurate tracking of eye movements using EyeLink 1000 Plus eye tracker (SR Research). The eye tracker was employed to measure interpreters' eye movement behaviours indicative of cognitive effort across different conditions, another key focus of our study. They filled out the NASA Task Load Index (Hart and Staveland, 1988) immediately after each task, to self-assess the workload they perceived using the NASA TLX iOS app. The eye-tracking data and self-assessment results will be analysed and reported in a future study. Each interpreting task lasted around 15 minutes on average. Their interpreting output was audio-recorded and transcribed verbatim for data analysis.

A retrospective session started after the completion of all tasks and a short break. Participants verbally reported their interactions with ASR in the recently performed interpreting tasks with some cues provided in a paper. The duration of this session varied between 7 and 19 minutes. In addition, participants partook in a semi-structured interview to share their overall attitudes towards using ASR in healthcare interpreting and offer suggestions for improvement. This part lasted between 19 to 49 minutes. The last two sessions were both audio-recorded and transcribed verbatim.

**Data analysis**

The interpreting quality was analysed using an adapted version of the NTR model (Romero-Fresco and Pöchhacker, 2017), an error-based framework originally developed to evaluate accuracy in interlingual subtitling. It has recently been adapted for assessing interpreting quality (e.g., Korybski *et al.*, 2022; Rodríguez González *et al.*, 2023) for its identification of translation errors.

The original NTR model consists of a formula and an overall assessment (Figure 3). In interpreter-mediated conversations, interlocutors usually only hear the interpreter's output. Recognition errors, therefore, did not apply to the interpreting workflow in this study and were not considered when using this formula. The formula calculates the accuracy rate, while the overall assessment comprises the accuracy rate, comments on issues not covered by the formula, such as effective editions, the speed, delay and overall flow of the interpreting output, and a final conclusion (Romero-Fresco and Pöchhacker, 2017, p.159). Ultimately, it is the overall assessment that indicates the quality.

The NTR model adopts a three-level grading system to classify errors by severity: "minor errors", "major errors" and "critical errors", deducting 0.25, 0.5 and 1 points respectively. As a meaning-focused model, it evaluates errors based on the "idea unit", defined by Chafe (1985) as a "unit of intonational and semantic closure", which typically encompasses a verb phrase along with a noun, prepositional or adverbial phrase. Minor errors cause largely insignificant deviations, major errors often result in isolated information loss, and critical errors produce an utterance with an entirely new meaning (Romero-Fresco and Pöchhacker, 2017, p.152).

$$\text{NTR} = \frac{\text{N-T-R}}{\text{N}} \times 100 = \%$$

Assessment

N: Number of target words

T: Translation errors

    content errors: omission, addition, substitution

    form errors: correctness (grammar and terminology) and style (appropriateness, naturalness, register)

R: Recognition errors (misrecognitions by the speech recognition software in respeaking-based live subtitling which viewers can see in subtitles)

EE: Effective editions (deviations from the source text that do not involve a loss of information or that even enhance the communicative effectiveness)

Figure 3: The NTR model (Romero-Fresco and Pöchhacker, 2017)

To conduct the assessment, all 16 interpreting outputs were transcribed verbatim into text, segmented into idea units and manually aligned with the source material in the NTR sheets. Although all participants performed bidirectional interpreting tasks, our current analysis addressed only the quality of English-to-Chinese interpreting. This focus was driven by the greater complexity of the doctor's utterances, which often included technical terms and numbers, coupled with the listening challenge of interpreting from a second language. To ensure evaluation consistency and reduce rating subjectivity, each output was analysed by two evaluators who received training on NTR evaluation before carrying out the task. When discrepancies arose, the evaluators engaged in discussions to reach an agreement.

Participants' retrospective reports were analysed to answer the third research question. Specifically, participants' reflections on the specific types of information they sought from ASR support, the way they used ASR and their preferences for the presentation of ASR output were examined.

## 4   Results

Table 4 presents the results of the interpreting quality assessment. Compared with the baseline condition (no ASR support), the average scores of interpreting quality under the three ASR-supported conditions all increased, by 0.52, 2.2 and 2.12 points respectively. Among the three conditions, interpreting with full ASR support yielded the highest mean score, while interpreting with partial ASR support had the lowest mean. The difference in interpreting quality between using full ASR transcripts and ChatGPT summaries was very small.

The breakdown of each participant's scores revealed that for three participants, their interpreting quality improved with ASR regardless of the types of ASR output and scored highest with full ASR support. Only Participant 4 had the lowest score with partial ASR support. The interpreting quality of Participant 1 and Participant 2 improved steadily as the amount of source text provided increased from partial ASR support to ASR-fed ChatGPT summaries, and finally to full ASR transcripts. In contrast, Participant 3 had her best performance with the ChatGPT summary.

Despite the limited sample size, an attempt was made to address the first two research questions by running inferential statistical tests using IBM SPSS Statistics (Version 26). Shapiro-Wilk test (Elliott and Woodward, 2007, p.25) confirmed that all data under each condition conformed to a normal distribution. One-way repeated measures ANOVA tests (Elliott and Woodward, 2007, p.175) were administered to all conditions. Mauchly's test of sphericity (Keppel and Wickens, 2004, p.376) was not violated ($p = .419 > .05$). A significant difference was found in interpreting quality across the four conditions, $F(3, 9) = 48.271$, $p = .000 < .01$, partial $\eta^2 = .942$. Therefore, post-hoc pairwise comparisons using Bonferroni correction ($\alpha = 0.05/6 = .0083$) (Elliott and Woodward, 2007, p.9) were conducted to find which pairs were significantly different (Table 5).

|  | C1 (no ASR) | C2 (partial ASR) | C3 (full ASR) | C4 (ASR-fed ChatGPT summary) |
|---|---|---|---|---|
| **P1** | 96.49 | 96.88 | 98.98 | 98.78 |
| **P2** | 96.63 | 97.15 | 98.44 | 98.29 |
| **P3** | 95.83 | 97.11 | 98.52 | 98.76 |
| **P4** | 97.45 | 97.34 | 99.27 | 99.03 |
| **Mean** | 96.60 | 97.12 | 98.80 | 98.72 |
| **Range** | 95.83-97.45 | 96.88-97.34 | 98.44-99.27 | 98.29-99.03 |
| **Standard Deviation** | .665 | .189 | .392 | .309 |

Table 4: Descriptive statistics for interpreting quality per participant under each condition

| Pairwise comparison (by condition) | Mean difference | Standard error | Sig. |
|---|---|---|---|
| 1 vs 2 | -.520 | .287 | .168 |
| 1 vs 3 | -2.203 | .227 | .002* |
| 1 vs 4 | -2.115 | .315 | .007* |
| 2 vs 3 | -1.682 | .197 | .003* |
| 2 vs 4 | -1.595 | .161 | .002* |
| 3 vs 4 | .087 | .111 | .487 |

Note: * for $p < .05$

Table 5: Results of post-hoc comparisons (repeated measures ANOVA)

To answer the first research question, the results revealed that the interpreting quality in the full ASR transcript condition ($M = 98.80$, $SD = .392$) and the ASR-fed ChatGPT summary condition ($M = 98.72$, $SD = .309$) was significantly higher than that in the condition without ASR ($M = 96.60$, $SD = .665$). However, no significant difference was observed between the partial ASR condition ($M = 97.12$, $SD = .189$) and the no ASR condition.

To answer the second research question, compared to the condition with partial ASR support, the interpreting quality was significantly higher with full ASR transcript and ASR-fed ChatGPT summary. There was no significant difference in the interpreting quality between the full ASR condition and the ChatGPT summary condition.

It should be noted that the power analysis showed a low statistical power of .141 (Elliott and Woodward, 2007, p.8), suggesting a limited capacity to detect true effects within the current sample. The inferential results may not be reliable due to insufficient power and therefore, should be interpreted with caution. However, a large effect size ($f = .728$) (Kelley and Preacher, 2012, p.147) was yielded by the sensitivity analysis using G*Power, indicating the substantial differences in interpreting quality across the conditions and the practical significance of the findings despite the low power.

As the NTR model allows us to access specific error types, the distribution of error types across various conditions (Figure 4) and per participant (Figure 5) was also examined.

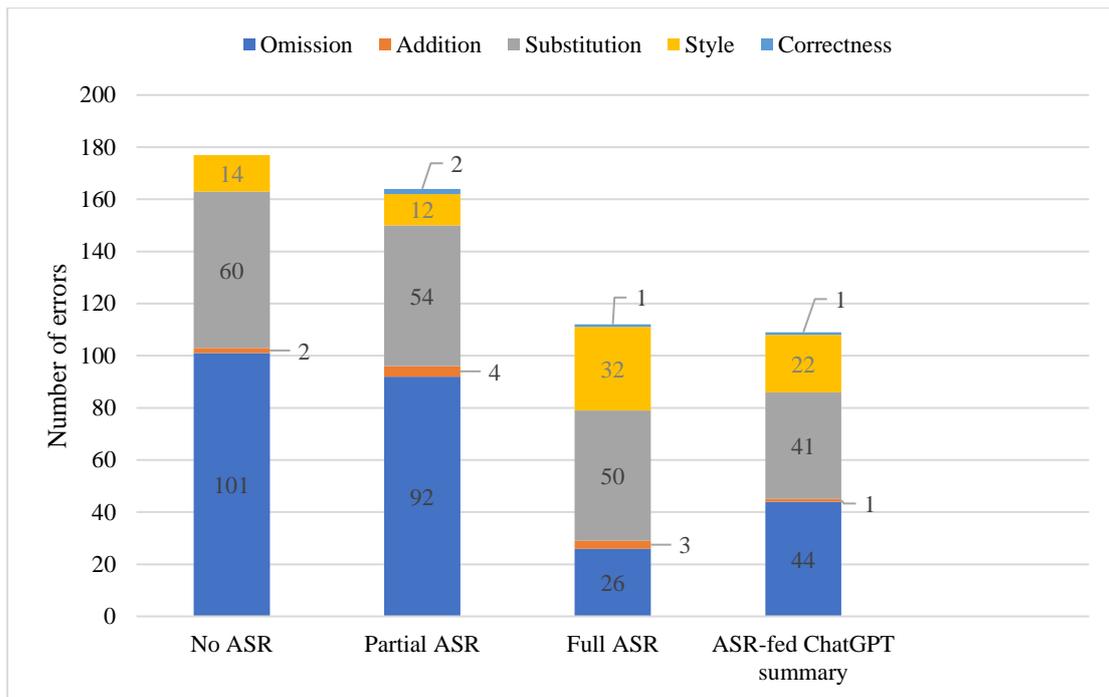

Figure 4: Error type distribution across conditions

Compared to the condition without ASR, the full ASR condition achieved the biggest reduction in omission errors, decreasing by 74.26%, followed by a 56.44% drop in the ChatGPT summary condition. However, the full ASR condition showed the highest increase in style errors, with the ChatGPT summary condition close behind. The ChatGPT summary condition witnessed the largest reduction in substitution errors, by 31.67%. The error type distribution between the no ASR condition and the partial ASR condition showed only modest differences.

The error distribution per participant under each condition is visualised in Figure 5. It shows that all participants made the fewest omission errors when assisted by full ASR transcripts, followed by ChatGPT summaries. Participant 2 was a major contributor to style errors observed in the full ASR condition (16 of 32 errors) and the ChatGPT summary condition (9 of 22 errors). This highlighted the need for a deeper examination of the errors made by individuals.

The third research question pertained to participants' interaction with ASR technology. Three participants shared that they relied on ASR support for medical terms, medicine names, dosages and units. One participant noted that she used ASR when she had difficulties understanding the interlocutor. When asked whether they used ASR support consistently throughout the task or only as needed, three participants believed that they primarily counted on their own listening skills and comprehension abilities, only resorting to the transcript when they struggled to understand the original utterance. One mentioned that after finishing interpreting an utterance, she occasionally reviewed the transcript to verify the accuracy of her delivery. Conversely, one admitted constantly using the provided transcript during interpreting and also expressed that the condition with only terms and numbers was distracting. However, preliminary analysis of interpreting errors (see Appendix A) and eye tracking data suggests that these participants frequently referred to ASR output during interpreting, evidenced by the reproduction of ASR errors in their interpreting output and patterns observed in eye fixation positions.

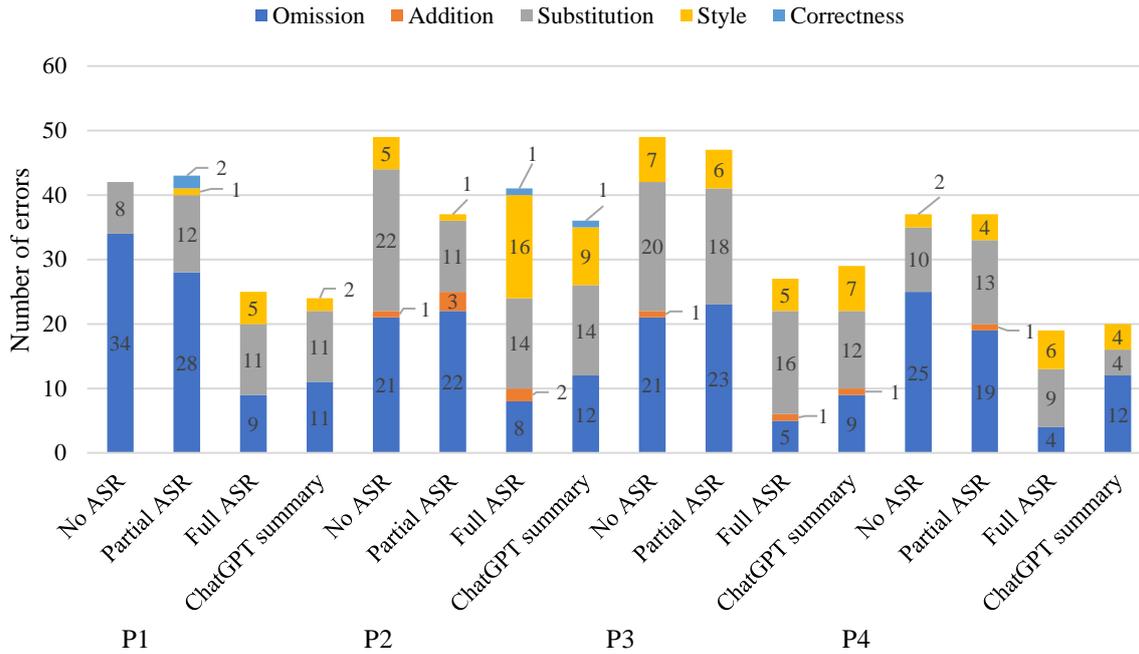

Figure 5: Error type distribution per participant

All participants expressed a preference for full ASR support if given the option. One explained that the information not provided by the partial ASR and ASR-fed ChatGPT summary conditions could be exactly what an interpreter might miss. Another participant mentioned that when full ASR output was available, she could allocate less effort to listening and just turn the task from interpreting into sight translation. Also, one would prefer to have the entire ASR transcript if the domain was foreign to her. However, she cautioned that in cases of familiar topics, the availability of a full ASR transcript could potentially interfere with interpreting by making interpreters overly dependent on it.

## 5 Discussion

The descriptive statistics indicated an increase in overall interpreting quality with varying types of ASR support. The integration of ASR into dialogue-based remote healthcare interpreting may have a positive impact on interpreting quality. Through inferential statistical analysis, we found that when the full ASR transcript or ASR-fed ChatGPT summary was provided, interpreting quality was enhanced significantly. This initial result was consistent with previous findings on the impact of captions on SI quality among student interpreters (e.g., Cheung and Li, 2022; Yuan and Wang, 2024) and professional interpreters (e.g., Rodríguez González *et al.*, 2023; Li and Chmiel, 2024). However, there was no significant difference in interpreting quality between the condition without ASR support and the partial ASR condition (numbers and terms). This contradicted previous research that revealed significant improvements in the accuracy of number renditions with numbers being provided (e.g., Desmet *et al.*, 2018; Defrancq and Fantinuoli, 2021; Pisani and Fantinuoli, 2021; Yuan and Wang, 2023). One explanation for this discrepancy is that previous studies primarily assessed the quality of numbers or terms only. Although our finding was embedded in the context of dialogue interpreting, it echoed Fritella's (2022) view that an ASR-integrated CAI tool may not necessarily facilitate number renditions in SI unless critical variables in the test speech were considered, for instance, the complexity of the speech, and a holistic assessment approach rather than concentration on the isolated numeral.

When examining the quality difference across varying ASR output conditions, the descriptive results showed that both the full ASR support and the ASR-fed ChatGPT summary were associated with a substantial improvement in interpreting quality, with only minimal differences detected between them. The provision of transcripts with only terms and numbers led to the smallest increase in mean interpreting quality scores. Inferential statistics suggested that interpreting quality was notably lower under the partial ASR condition compared to other ASR conditions, while no statistical difference in interpreting quality was found between the full ASR and the ChatGPT summary conditions. This initial finding implied that both full ASR transcripts and ChatGPT summaries were effective in improving interpreting quality. However, it did not suggest that partial ASR support should be ruled out when offering ASR solutions to interpreters. This result may be attributed to participants' limited experience in integrating ASR into healthcare interpreting. Professional interpreters may benefit from this type of ASR output as it may serve as a solution for "problem triggers" and reduce superfluous distraction. Given the lack of previous studies comparing the impact of varying types of ASR output, this finding provided a starting point for exploring the optimal ASR presentation that benefits interpreters most.

When taking a closer look at the error type distribution, it was found that the availability of full ASR transcripts helped all participants deliver more complete renditions by effectively addressing omission errors. Although nearly half of the stylistic errors in both the full ASR and the ChatGPT summary conditions were attributable to one participant, the high frequency of style-related errors in both conditions may suggest an association between these ASR outputs and less satisfactory stylistic appropriateness in interpreting. This issue was mainly manifested as disfluency phenomena, including filled fillers and silent pauses in participants' delivery which was consistent with Rodríguez González *et al.*'s (2023) findings in ASR-supported remote SI.

Regarding the interaction between interpreters and the technology, participants reported how they applied ASR output to the tasks and their preferences for ASR output presentation. They reported selectively using ASR support, mainly for specialised terms and comprehension issues. Three participants believed that they relied more on themselves than ASR. All participants preferred having access to a full ASR transcript, mentioning benefits such as improved completeness of information delivery, reduction of listening and analysis demands, and assistance with unfamiliar topics. However, as these results were based on self-reports from a sample of four, they should be interpreted with prudence.

## 6   Conclusions and future plans

In this pilot study, we explored the impact of ASR on remote dialogue interpreting in healthcare settings. The preliminary findings suggested that the availability of full ASR transcripts or ASR-fed ChatGPT summaries improved interpreting quality. However, access to transcripts of numbers and terms did not contribute to better interpreting quality. Participants' self-reported interactions with ASR were generally consistent, including the selective use of ASR output and a preference for full ASR transcripts.

These findings should be treated with caution, as they were exploratory and based on a very limited sample size. A number of limitations should also be acknowledged. First, the findings only reflected trainee interpreters' performance and experience with ASR support. They cannot be generalised to professional interpreters who will be the focus in our main study. Second, to ensure accurate eye tracking, note-taking was prohibited during the experiments. This may have had an impact on the interpreting quality, especially in the condition where no ASR was provided. To minimise the impact of confounding variables, common turn-taking issues, such as overlapping speech and interruptions in interpreter-mediated conversations were avoided in the simulated consultations. Moreover, the video remote interpreting interface with ASR support designed by our research team may be less familiar to the participants than those common commercial platforms with captioning features. These factors posed a risk of reducing the study's ecological validity. Third, as this study only evaluated English-to-Chinese interpreting output, these findings may not fully represent the quality of the entire bidirectional dialogue interpreting. Finally, given the observed limitations in applying the NTR model to assess healthcare interpreting quality, modifications may be necessary to adapt the model more effectively to this interpreting scenario.

To conclude, this pilot study successfully validated the methodology. In the next phase of our study, we will analyse the eye-tracking data to investigate how participants allocated their cognitive effort when different types of ASR output were provided. A detailed comparison between ASR transcription errors and interpreting errors will be performed to explore how they may have used ASR support. We will also refine our research design and conduct the main study with a larger sample of interpreters who are experienced in healthcare interpreting.

Yuan, Lu, and Binhua Wang. 2024. Eye-tracking the processing of visual input of live transcripts in remote simultaneous interpreting: Preliminary findings. *FORUM*, 22: 118-144.

## Appendix A: Examples of Interpreting Errors

**Condition 1: No ASR support**

| Original script | ASR output | Interpreting output | Back translation |
|---|---|---|---|
| **Excellent, that's important for managing stomach acidity effectively.** | N/A | 很好,这非常重要。 | Great, that's very important. |
| Error type(s): omission error<br>Analysis: The interpreter left out the reference to stomach acidity management. | | | |

**Condition 2: Partial ASR support**

| Original script | ASR output | Interpreting output | Back translation |
|---|---|---|---|
| **And your serum creatinine has decreased to 1.6. Your urea levels are also better at 50 down from 70 last check.** | serum creatinine - 血清肌酐<br>urea levels - 尿素水平<br>1.6 - 1.6<br>50 - 50<br>70 - 70 | 还有这个血清肌酐以及呃这个尿素水平也是…是 1.6, 50 和 70 这个数据。 | And your serum creatinine and um urea levels are also…are 1.6, 50 and 70. |
| Error type(s): omission error<br>Analysis: The interpreter omitted phrases indicating changes and instead only stated the numbers. | | | |

**Condition 3: Full ASR support**

| Original script | ASR output | Interpreting output | Back translation |
|---|---|---|---|
| **To reduce the inflammation, you will start with a high dose of corticosteroids, specifically Prednisone at 60 mg daily.** | To reduce the inflammation, you will start with a high dose of corticosteroids [皮质类固醇], specifically Prednisone [泼尼松] at 60 minutes daily. | 要想减少炎症,首先呃你要服用高剂量的皮类…皮质类固醇,尤呃尤其是泼尼松,大概每天 60 分钟。 | To reduce the inflammation, first um you will start with a high dose of cor…corticosteroids, esp um especially Prednisone, for about 60 minutes per day. |
| Error type(s): substitution error<br>Analysis: The interpreter followed ASR's transcription error. | | | |
| **I can see that you don't have oedema, your chest is clear, and your abdomen is soft and not tender.** | I can see that you don't have. Your chest is clear and your abdomen is soft and not tender [一碰就痛]. | 呃我…我看到了你没有过敏原。你的…呃胸部是…胸腔是很干净的,你的腹部很柔软,并没有一碰就痛。 | Um I…I can see you don't have allergies. Your…um your breast is…chest is clear, and your abdomen is soft and not tender. |
| Error type(s): substitution error<br>Analysis: ASR omitted the term "oedema," which the interpreter then incorrectly substituted with "allergies." | | | |
| **From the test results, it looks like your diabetes has not been well controlled, which could be contributing to your symptoms.** | From the test results, it looks like your diabetes has not been well controlled, which could be contributed to your symptoms. | 呃根据检查结果,呃看起来你的糖尿病呃已经得到了很好的控制,呃这些可以导致你现在的这个状况…症状的。 | Um from the test results, um it looks like your diabetes has been well controlled, um which could be contributing to your current situation... symptoms. |
| Error type(s): substitution and style errors | | | |
| **Analysis:** The ASR transcription was correct, but the interpreter reversed the meaning and frequently used filled fillers like "um." | | | |
| **You also have a congestive heart failure and ischemic heart disease, which are affecting your circulation.** | You also have a congested portfolio and ischemic heart disease [缺血性心脏病], which are affecting your circulation. | 呃同时你有去做一些检查以及心脏的缺血…有缺血性心脏病，这些也有可能会影响到你的循环。 | Um at the same time you did some checks and the ischemia in the heart … there is ischemic heart disease, which might also affect your circulation. |
| **Error type(s):** addition and omission errors<br>**Analysis:** The interpreter added, "Um at the same time you did some checks." ASR mis-transcribed "a congestive failure" as "a congestive portfolio," but the interpreter did not follow this error and instead omitted it. | | | |

**Condition 4: ASR-fed ChatGPT summary**

| Original script | ASR output | Interpreting output | Back translation |
|---|---|---|---|
| **To reduce the inflammation, you will start with a high dose of corticosteroids, specifically Prednisone at 60 mg daily.** | • To reduce inflammation, start with a high dose of corticosteroids [皮质类固醇], Prednisone [泼尼松] at 60 mg daily [每天 60 毫克]. | 我们呃现在是需要呃解决你，减少你这个炎症，我们是需要用到类固醇，呃还有泼尼松，每天需要有 60 毫克。 | We um currently need to um solve you, reduce your inflammation. We will use steroids, um also Prednisone at 60 mg daily. |
| **Error type(s):** omission error<br>**Reason:** The ChatGPT summary corrected the ASR transcription error by changing "60 minutes" to "60 mg," which helped avoid any misunderstanding for the interpreter. Here is only a minor omission of the phrase "a high dose of." | | | |
| **Yes, your test results indicate you have anaemia, diabetic nephropathy, hypertension, a history of myocardial infarction and stroke.** | • Test results indicate:<br>Anaemia [贫血]<br>Diabetic nephropathy [糖尿病肾病]<br>Hypertension [高血压]<br>History of myocardial infarction [心肌梗死]<br>Stroke | 嗯，[哎]…是的，你的检查结果显示你有贫血，还有这个糖尿病，你高血压、心肌梗死、中风。 | Um, [sigh]…yes, your test results indicate you have anaemia, also diabetes, you hypertension, myocardial infarction and stroke. |
| **Error type(s):** substitution and style errors<br>**Reason:** The interpreter incorrectly substituted "diabetic nephropathy" with "diabetes." The style error was noted due to the interpreter's use of fillers like "um" and even a sigh, along with the listing of conditions by repeating the ChatGPT summary verbatim without any transitional words, as in "you hypertension … ." | | | |